\documentclass[letterpaper, 10 pt, conference]{ieeeconf}  

\IEEEoverridecommandlockouts                              

\usepackage{times}

\usepackage{algorithm}
\usepackage[noend]{algcompatible}

\usepackage{amsmath}
\usepackage{amsfonts}
\usepackage{breqn}
\usepackage{multirow,hhline,graphicx,array}
\graphicspath{ {./images/} }
\usepackage{multicol}
\usepackage[bookmarks=true]{hyperref}
\usepackage{subfiles}
\usepackage{caption}
\usepackage{float}
\usepackage[flushleft]{threeparttable}

\usepackage{xcolor}

\title{\LARGE \bf
Predict the Rover Mobility over Soft Terrain using Articulated Wheeled Bevameter 
}

\author{	Wenyao Zhang,  
	Shipeng Lyu,	
	Feng Xue,
	Chen Yao,
	Zhengtao Liu,
	Zheng Zhu,
    and Zhenzhong Jia*,
\thanks{All authors are with the Shenzhen Key Laboratory of Biomimetic Robotics and Intelligent Systems, Department of Mechanical and Energy Engineering, Southern University of Science and Technology (SUSTech), Shenzhen, 518055, China.
They are also with Guangdong Provincial Key Laboratory of Human-Augmentation and Rehabilitation Robotics in Universities, SUSTech, Shenzhen, 518055, China.
*Corresponding author: {\tt\small jiazz@sustech.edu.cn}}%
}

\begin{document}

\maketitle
\thispagestyle{empty}
\pagestyle{empty}


\label{sec:abst}

\begin{abstract}

Robot mobility is critical for mission success, especially in soft or deformable terrains, where the complex wheel-soil interaction mechanics often leads to excessive wheel slip and sinkage, causing the eventual mission failure.
To improve the success rate,
online mobility prediction using vision, infrared imaging, or model-based stochastic methods have been used in the literature.
This paper proposes an on-board mobility prediction approach using an articulated wheeled bevameter that consists of a force-controlled arm and an instrumented bevameter (with force and vision sensors) as its end-effector.
The proposed bevameter, which emulates the traditional terramechanics tests such as pressure-sinkage and shear experiments, can measure contact parameters ahead of the rover's body in real-time,  and predict the slip and sinkage of supporting wheels over the probed region.
Based on the predicted mobility, the rover can select a safer path in order to avoid dangerous regions such as those covered with quicksand. 
Compared to the literature, our proposed method can avoid the complicated terramechanics modeling and time-consuming stochastic prediction; it can also mitigate the inaccuracy issues arising in non-contact vision-based methods.
We also conduct multiple experiments to validate the proposed approach.


\end{abstract}


\section{Introduction}
\label{sec:Intro}

%
%
%

Over the past several years, the working scene of autonomous robots has extended from indoor to off-road field environments and even on other planets. 
Compared to the on-road environment, field environments are challenging for ground robots to navigate over because the terrain and obstacle types are complex and diverse, and a series of characteristics such as the roughness and slope of the ground will also cause difficulties. 
In most unstructured terrains, robots can use lidar and cameras to finish the perception tasks, but soft terrains contain a variety of materials and complex terramecanical properties, which brings huge difficulties to the robot's mobility, especially for wheeled robots \cite{shrivastava2020material}.
In order to achieve autonomous operations such as resource exploration and planetary exploration, robots need to have more intelligent environmental perception and understanding capabilities to predict quickly and accurately mobility.

\begin{figure}[ht]
    \centering
    \includegraphics[width=\columnwidth]{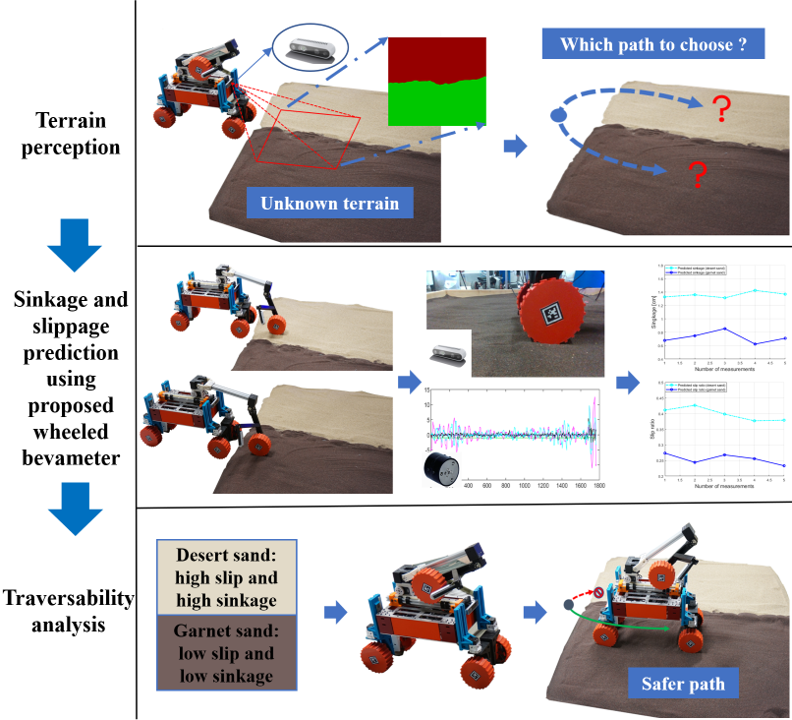} 
    \caption{Illustration of our pipeline for the mobility prediction over soft terrain using our proposed wheeled bevameter. 
    The robot can only detect the geometry and color information of the unknown terrain using the on-board vision sensor.  However, it cannot tell the physical properties of the terrain, i.e., it cannot differentiate the soft loose soil from the firm soil without making physical contact.  Hence, the robot is unable to choose the safer path.
    To solve this problem, the robot uses the wheeled bevameter to probe the unknown region by measuring the sinkage and slippage of the bevameter wheel using the F/T sensor and camera.
    The robot can then predict the physical properties of the terrain and the mobility of the supporting wheels of the probed region, thereby selecting a safer path with low sinkage and slippage.
    }
    \label{fig:whole_robot_model}
\end{figure}

Soft terrain has fluid-solid duality and complex terramicanical properties \cite{naclerio2021controlling}.
Therefore, the robot will often face excessive and abnormal slippage and sinkage on soft terrain.
When the slip ratio is too large, the robot will be unable to move and dig pits on the spot, resulting in an extremely large increase in sinkage and even damage to the robot \cite{kerr2009mars}.
As a result, we should focus on key factors like sinkage and slip ratio about soft soil to predict and analyze the traversability accurately.
Predicting the sinkage and slip ratio without danger is an important capability on soft terrain in a field environment.

The perception methods of unstructured terrain are currently not perfect, robots can only simply avoid obstacles and classify terrain, which is not enough for the off-road robots.
Proprioceptive-based methods based use IMU (Inertial measurement unit), force sensor, or other proprioceptive sensors can only know the type of terrain where the robot is touching, and the robot may be already in danger when it finds anomalies.
Remote measurement using sensors like lidar and camera may classify terrains wrongly with similar geometrical characteristics. 
Additionally, surfaces with the same material may have totally different results as the size of granule or moisture. Although the hybrid method can obtain higher measurement accuracy, it can not predict parameters about traversability such as the sinkage and slip ratio.
Meanwhile, the data we get from indoor experiments can't give exact information about the wild environment. 
And the traditional soil-wheel interaction model has larger errors for robots with small wheels.

%

We propose wheeled bevameter as a solution to the problems above. 
We adopt a force-control mechanical arm with a wheel to actively explore the area in front of the rover.
And use computer vision and tactile to measure sinkage of the wheel and slip ratio  to predict the traversability of the sandy region.


This paper has two main contributions as follow:

1. A robust and accurate method of vision-based wheel-soil contact parameters estimation.

2. Building unique articulated wheeled bevameter to predict sinakge and slip ratio of unknown terrain, analyze mobility and choose a safer path without complex model-based calculation and traveling on the unknown and dangerous terrain.

Subsequently, we describe the related work in Sec. II and introduce our hardware in Sec. III. 
Then we describe our methods of estimating contact surface in Sec. IV. and predicting sinkage and slip ratio in Sec. V.
We apply those to our robot and do traversability analysis in Sec. VI. Finally, we conclude the work in Sec. VII. 
\section{Related Work}
\label{sec:Related_work}

\subsection{Terrain Perception}
\label{sec:Terrain-Detection}
To gather terrain information, there are three perception methods: proprioceptive-based methods, exteroceptive-based methods, and hybrid methods.
Indoor tests, such as the single-wheel testbed experiment, can also yield terramecanics parameters, which can be used for soft terrain navigation.
All the methods above aim to get a lot of information about surrounding terrain to predict the traversability of robots or choose the best path to travel.

Proprioceptive-based methods primarily use vibration or accelerometer signals collected by IMU and tactile sensors to classify the terrain to predict traversability and choose safer terrain.
Early proprioceptive-based measurement originated from planet rover. 
Ref.\cite{2002Terrain,2005Vibration} classify the terrain type via vibration and acceleration data from IMU.
\cite{2006Vibration} process the accelerate signal and compare the effect of six different SVM (Support Vector Machine) in advance. 
\cite{libby2019self} and \cite{valada2017deep} tried to classify the ground using audio sensors, but it could not be used in a space or noisy environment.
\cite{MA2010Haptic} combine motor current signal with IMU signals of the legged robot to classify ground, and \cite{8630524} tried to combined force sensors with IMU signal.
In addition, \cite{xu2017measurement} measure the wheel-terrain contact angle for robot on rough terrain using a laser scanning sensor.
Because these methods require the robot to tour the region in order to make predictions, their applications are constrained to safe and predictable environments and do not work well in unknown and hazardous environments.


Lidar and camera can measure the ground remotely and are widely used in terrain perception these years.
\cite{helmick2009terrain},\cite{ono2015risk},\cite{siva2019robot} classify the terrain type according to geometry, texture and color information based on classification algorithm.
With the development of deep learning, semantic segmentation are used gradually to identify ground and estimate some terramechanics parameters. (\cite{rothrock2016spoc},\cite{maturana2018real},\cite{2019Mapping},\cite{iwashita2019mu}).
But there are comparatively fewer off-road datasets than urban environments, so this method may not be used widely now.
However, exteroceptive-based methods light, rain and fog will affect our perception accuracy of exteroceptive-based methods.

Given the complimentary outputs of different perception methods, there are also hybrid methods that take advantage of the strengths of each perception method to predict traversability.
\cite{halatci2007terrain},\cite{2012Self},\cite{2016jplvisionandtandactile},\cite{2021FreiburgTRO} combined many signals such as IMU, camera, audio sensors and adopted self-training approach to reduce the label cost and improve accuracy of classification.


To estimate sinkage and slip ratio in the past, we had to integrate terramechanics parameters with a sophisticated model.
The gathering of terramechanics parameters, on the other hand, necessitates enormous amounts of experimental data and bulky experimental equipment.
Numerous studies have attempted to set up the single-wheel platform and improve the model based on Wong's wheel-soil interaction model\cite{wong1967prediction}.
for example,\cite{yoshida1996development},\cite{apostolopoulos2001analytical},\cite{center2004rover},\cite{iagnemma2005laboratory},\cite{ding2011experimental}. 
\cite{jia2012terramechanics},\cite{ding2015identifying} proposed some methods for analyzing model parameters more quickly.
Single-wheel testbed, however, is too huge to be mounted on robots and cannot be used for real-time prediction.

\subsection{Traversability Analysis}
The goal of parameters estimation and terrain classification is to predict traversability and ensure the safe execution of tasks.
\cite{2010Predictable} based on stochastic response surface method and uncertainty to analyze.
\cite{wermelinger2016navigation} predicted the traversability combine slope, roughness and step height with dynamics model of legged robot, and created a traversability map.
\cite{fan2021step} predicted the traversability over unknown terrain based on collision, step, slippage, and tip over.
However, these studies did not consider the sinkage and slippage of soft ground.


Based on the researches above, we design an unique and active detection tool to explore terrain ahead actively. 
We may obtain the sinkage and slip ratio information to estimate traversability and choose the safer path without having to calculate complex mechanic models.

\section{System Setup}
\label{sec:Hardware}


%


To validate the mobility prediction approach outlined in Fig.\ref{fig:whole_robot_model}, we build an experiment platform consisting of a robot base with independent suspension, an articulated wheeled bevameter, and a sand box as the test field, as shown in Fig.\ref{fig:robot platform}.
The articulated wheeled bevameter can be further divided into a robot arm (capable of force control) and an instrumented bevameter as its end-effector.
The F/T sensor and camera installed on the bevameter can measure the contact forces, sinkage, slippage, and wheel-soil interaction details.
Hence, the robot can use the bevameter to explore the unknown terrain ahead, without risking of moving the supporting wheels into the unexplored risky regions such as those covered with quicksand where the robot can easily get trapped in.
Moreover, our mobility prediction method does not require complicated terramechanics modeling, thereby ensuring prediction accuracy by eliminating errors caused by terrain parameter estimation and wheel-soil interaction modeling, which can be found in other methods such as \cite{2010Predictable}. 


\subsection{Rover Base with Independent Suspension}
\label{sub:Newtons_law}

\begin{figure}[ht]
    \centering
    \includegraphics[width= \columnwidth]{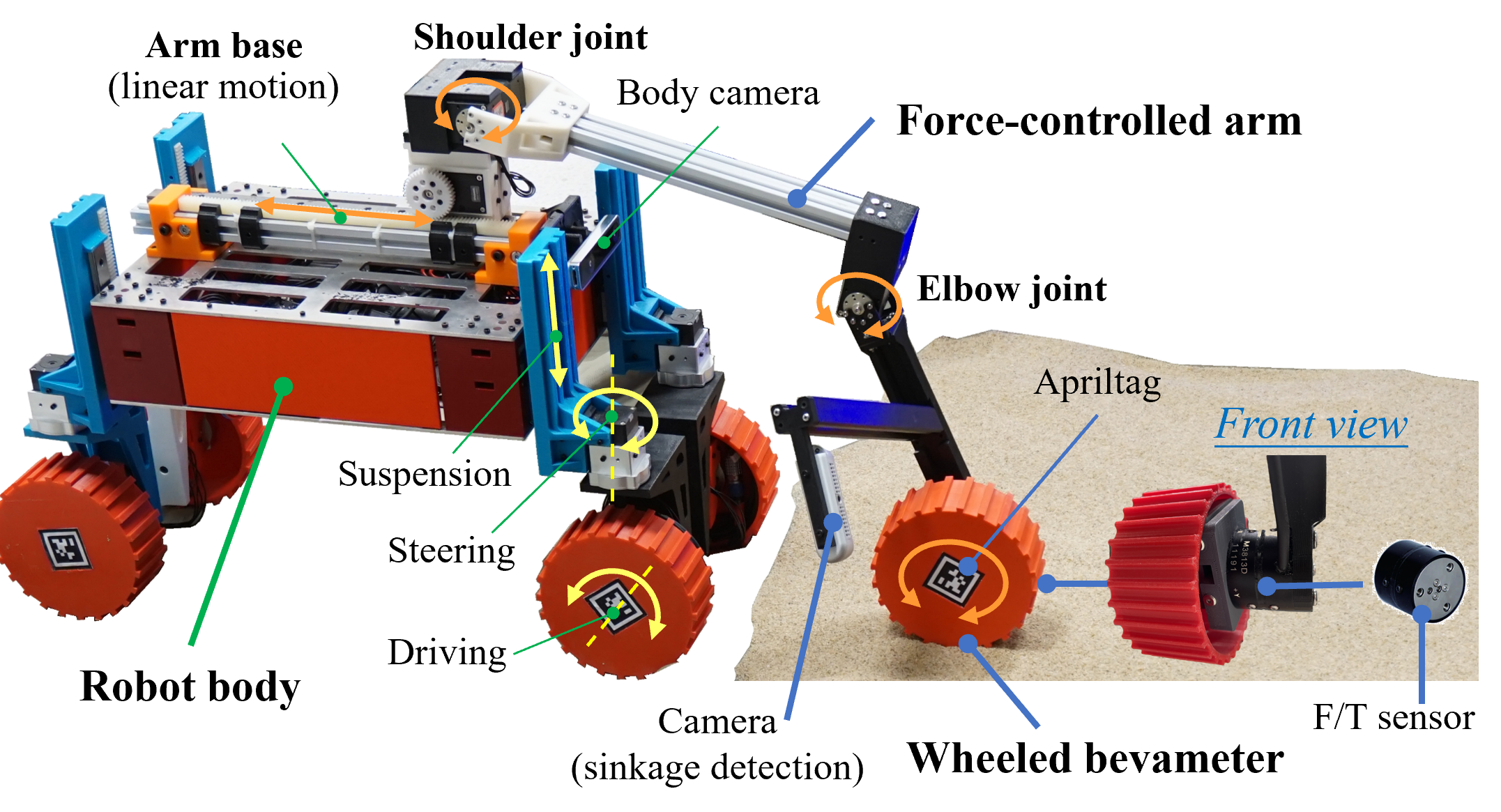}
    \caption{The robot platform has a 4WD-4WS configuration with active suspension.  
    The robot is equipped with an articulated force-controlled arm with a wheeled bevameter (has F/T and vision sensors) as its end-effector to explore the unknown terrain in front of the robot, in order to better predict the mobility of robot driving wheels over the probed region.
    }
    \label{fig:robot platform}
\end{figure}

As shown in Fig.~\ref{fig:robot platform}, our four-wheel-drive and four-wheel steering (4WD-4WS) rover base has individual articulated suspension to adjust the robot pose.
The rover base is about $60cm$ long, $45cm$ wide and its height can be adjusted in the range of $35-45cm$.
The rover base is equipped with IMU (inertial measurement unit) and stereo vision camera (body camera in Fig.~\ref{fig:robot platform}) so that it can monitor its status such as velocity and position while perceiving its surrounding environment.
Other vision system such as RGB-D cameras can also be installed on the rover base to capture unstructured terrain information for SLAM and navigation purposes.
For experimental validation purposes, We install F/T sensors (model: SRI M3813D) on each driving wheel to compare the forces acting on the driving wheel to those of the bevameter wheel.
The robot can be operated at autonomous mode or teleoperated through a joystick or a computer.

%

\subsection{Articulated Wheeled Bevameter}
\label{sub:Bevameter}

As shown in Fig.~\ref{fig:robot platform}, one innovation of our paper is that we use articulated wheeled bevameter, which can be further divided into an articulated arm and an instrumented bevameter as the arm's end-effector, to explore the physical properties of the unknown region and predict the mobility of robot driving wheels over the probed area.

As shown in Fig.~\ref{fig:robot platform}, the arm has three joints, namely: the arm base consisting of a servo and a linear stage, the shoulder joint driven by two servos connected in parallel, and the elbow joint.
The arm can also have other configurations.  For example, we can add a waist joint to increase the arm's workspace.
The end-effector of the arm is an instrumented wheeled bevameter consisting of a wheel, a servo used to drive the wheel, and a 6-axis F/T sensor (model: SRI M3813D).
For rapid prototyping purposes, we use aluminum extrusion profiles, 3D-printed parts, and $5$ high quality Dynamixel XH servos capable of torque control to build the articulated wheeled bevameter.  Its entire weight is $2.5kg$.
We have achieved hybrid position and force control using these servo motors and the F/T sensor.
During experiments, we can vary the vertical loading forces of the bevameter wheel, and its traveling speed and slip ratios, and measure the traction forces and driving torques, thereby executing standard terramechanics tests such as the pressure-sinkage and shear experiments to estimate the terrain properties, as explained in Sec.\ref{sec:bevameter-experiment}.

%

We mount a camera (Intel D435i) on the arm and an Apriltag on the side surface of the wheel to robustly detect multiple wheel-soil contact parameters such as sinkage and contact angles, as well as recording experiment details, as shown in Sec.\ref{sec:Vision-based-est}.
The recorded wheel-soil interaction process can be used in terramechanics modeling or parameter analysis.
We develop algorithms that are robust against complex background, view-point and illumination changes for vision-based contact geometry estimation, as illustrated in Sec.\ref{sec:Vision-based-est}.
The camera's position is quite important.  There might be occlusion problems if the camera directly points to the wheel's side surface, i.e., the camera's optical axis aligns with the wheel's rotation axis.  After many tests, we place the camera diagonally above the wheel to get a better view.
Note that the camera placement in Fig.\ref{fig:robot platform} is mainly for feasibility test. 
We can freely adjust the camera position because our algorithm is robust against view-point variations thanks to the built-in self-calibration function.
We can even use space-saving foldable designs for practical considerations during the actual deployment.

In our setup, we prefer to use a bevameter wheel that is the same as the driving wheel in every perspective such as geometry and wheel surface patterns.
One advantage of this is that we can directly calculate the forces, sinkage and slippage of the driving wheel, without going through the complicated terramechanics modeling process.
This helps to eliminate terrain parameter estimation error and terramechanics modeling error, both of which can significantly affect the prediction accuracy \cite{2010Predictable}.
We can also use a down-scaled bevameter wheel (to save some space) and estimate the sinkage and slippage of the driving wheel using simple calculations. This has been experimentally validated.
%

\subsection{Other Applications of the Wheeled Bevameter}
\label{sub:bevamter-application}

\begin{figure}[ht]
    \centering
    \includegraphics[width=\columnwidth]{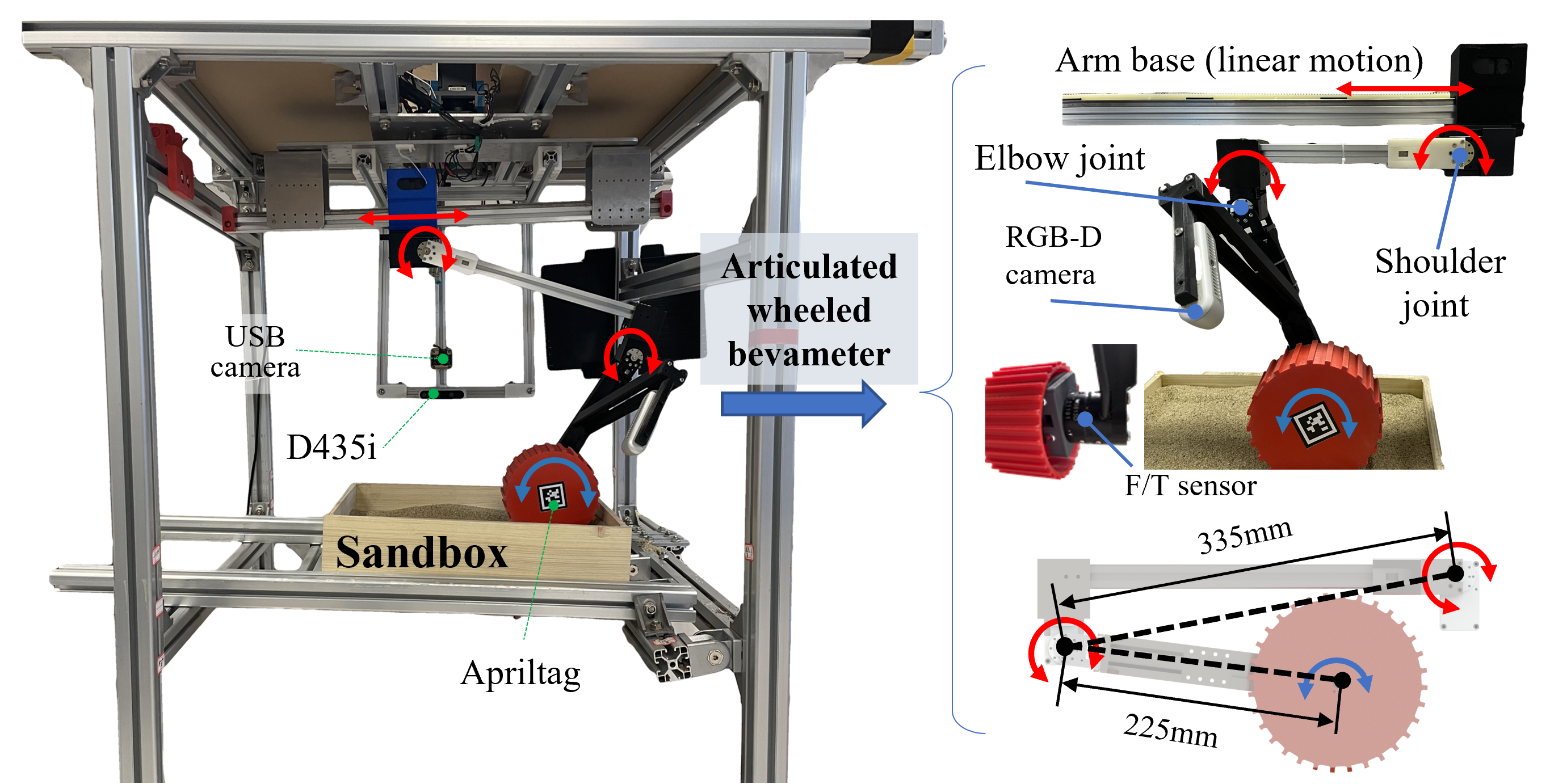}
    \caption{The proposed articulated wheeled bevameter can be reconfigured into a stationary testbench for wheel-soil interaction studies.}
    \label{fig:bevameter-reconfig}
\end{figure}

Besides probing the unknown terrain properties, our proposed wheeled bevameter can be used to restructure or modify the surrounding soils for improved mobility
\cite{shrivastava2020material}. 
It can also be reconfigured into a stationary testbench for wheel-soil interaction studies, as shown in Fig.~\ref{fig:bevameter-reconfig}.
Using this equipment, we can conduct traditional terramechanics tests such as the pressure-sinkage and shear experiments, as shown in Sec.~\ref{sec:bevameter-experiment}.

\subsection{Soils Investigated in This Study}
\label{sub:Soils}

Granular terrain is a typical medium covered on the planetary surface, especially Mars and Moon surfaces. This terrain
is generally composed of gravel, dust, and other materials.
To simulate the mechanical properties of the real exploration environment, we can complete experiments with multiple mediums by changing the kinds of the granular medium. 
We change the level and loosen of terrain manually to ensure the uniformity of the medium in each experiment.

\begin{figure}[H]
\label{soil_figure}
\centering
\includegraphics[width=\columnwidth]{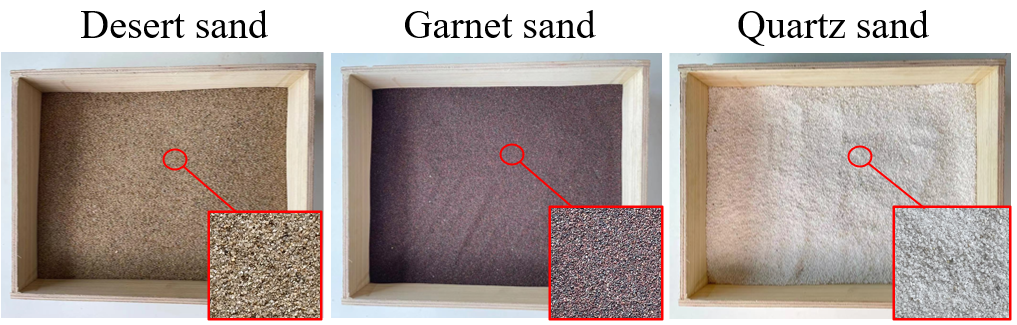}
\caption{Three types of granular medium (sand) used for experiments.}
\label{fig:soil}
\end{figure}

The most common granular medium terrain in nature are deserts, so we choose the sand in the Tengger Desert as an experimental medium.
This sand's granularity is not uniform, and its grit mesh number is between 30 and 60, which can represent the most real environments with granular terrain. 
In order to study the influence of different sandiness and grit thickness (mesh) on the contact event, we also choose the other two kinds of sand. One is garment sand with 60 meshes, the other one is quartz sand with 30 meshes. 
The derail medium is as shown in Fig.\ref{fig:soil}.

\section{Vision-based wheel-soil contact geometry estimation}
\label{sec:Vision-based-est}
In this section, we propose a method of contact parameter estimation based on vision, and the experimental results prove that our algorithm has great robustness under different light conditions and complex backgrounds.
Meanwhile, our algorithm also has a high accuracy and detection rate with low latency.

%
%


\begin{figure}[ht] 
\centering 
\includegraphics[width= \columnwidth]{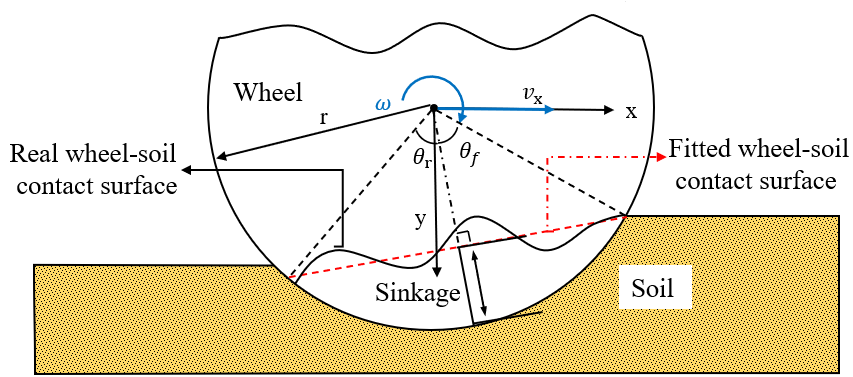} 
\caption{Sinkage definition on soft terrain.} 
\label{fig:sinkage_defination} 
\end{figure}

When driving across soft soils, such as sand, loose dirt, or snow, robots will sink. 
Multiple factors contribute to this phenomenon such as load, surface material, granule size, moisture. 
Ref.\cite{2017Sinkage} define sinkage on soft terrain as the Fig.\ref{fig:sinkage_defination}.
In this figure, $r$ is defined to be the radius of the wheel, $w$ is the rotation speed, $\theta_f$, and $\theta_r$ denote the entry contact angle and the exit angle, respectively.
Sinkage is an important parameter in wheel-soil interaction and can be applied to analyze terramechanics properties parameters and traversability. 
So it is important to measure accurately sinkage when robots move on soft terrain.

\begin{figure}[ht]
\centering
\includegraphics[width=\columnwidth]{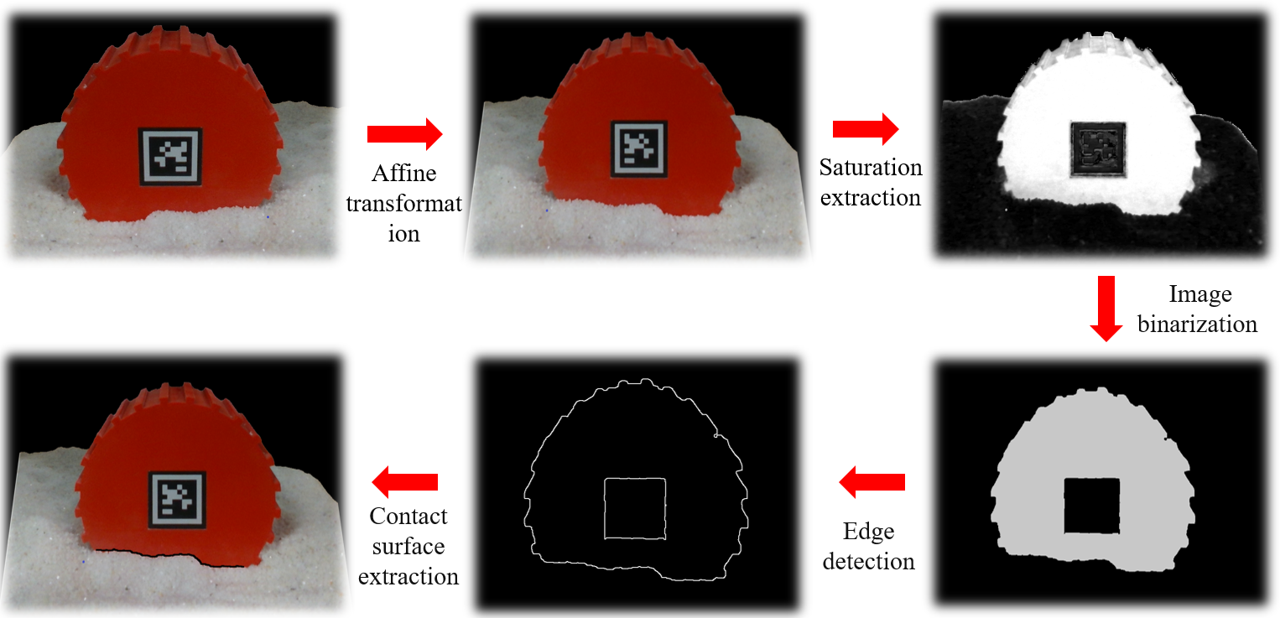}
\caption{Pipeline of contact parameter estimation.}
\label{fig:sinkage-detect-pipline}
\end{figure}

In wheel-soil parameter estimation based on vision, the most important two parts are to distinguish the wheel-soil surface and locate the center of the wheel.
To extract the boundary of the wheel and the ground, the edge of the wheel must be detected.
Hough transform \cite{ballard1981generalizing} can be used to extract rim, but this algorithm relies on parameter adjustment too much and fail to detect the wheel when part of the wheel is outside the camera vision.
We provide details of the experimental procedures carried out during this investigation as Fig.~\ref{fig:sinkage-detect-pipline}.

Now we discuss each step in detail:
\begin{itemize}
    \item \emph{\textbf{Affine transformation:}} the camera plane is not parallel to the wheel plane, we need to make a affine transformation on the original RGB image as formula\ref{eq: transform-formula} to make sure four corners of Apriltag in the same plane. 
    \begin{equation}
     F_{n}=\left[ \begin{array}{c}
    	R \quad T \\
    	0 \quad 1 \\
    \end{array} \right] F_{i}
    \label{eq: transform-formula}
    \end{equation}
    $R$ is a 3*3 rotate matrices, $T$ is a 3*1 translate matrix. $F_i$, $F_n$ denote initial image and the processed image.

    \item \emph{\textbf{Saturation extraction:}} RGB information is greatly affected by brightness.
    On the contrary, saturation is independent of brightness, so we can reduce the effect of the illumination condition to get a better result.	
    We transform the image from RGB space to HIS space to get saturation using the following formula.

    \begin{equation}
    \left\{
    \begin{gathered}
        I = \frac{R+G+B}{3}  
        \\[3pt]
        S = 1 - \frac{min(R,G,B)}{I}
        \end{gathered}
      \right.
    \end{equation}


    \item \emph{\textbf{Image binarization:}} Apply OTSU\cite{Otsu1979ATS} algorithm to image binarization, then use opening operation algorithm to delete small parts and fill the holes (noise created by sand). 
    
    \item \emph{\textbf{Edge detection:}} Use Canny operator\cite{canny1986computational} to detect edge and Apriltag detection algorithm to locate the center.
    According to the center position and radius of the wheel, we can extract the wheel-soil interaction interface.
    
    \item \emph{\textbf{Contact surface extraction:}} Fit the wheel-soil interaction surface with robust regression\cite{rousseeuw2005robust}, then calculate the wheel sinkage, entrance, and exit angle.
 
\end{itemize}

\begin{figure}[ht] 
\centering 
\includegraphics[width=\columnwidth]{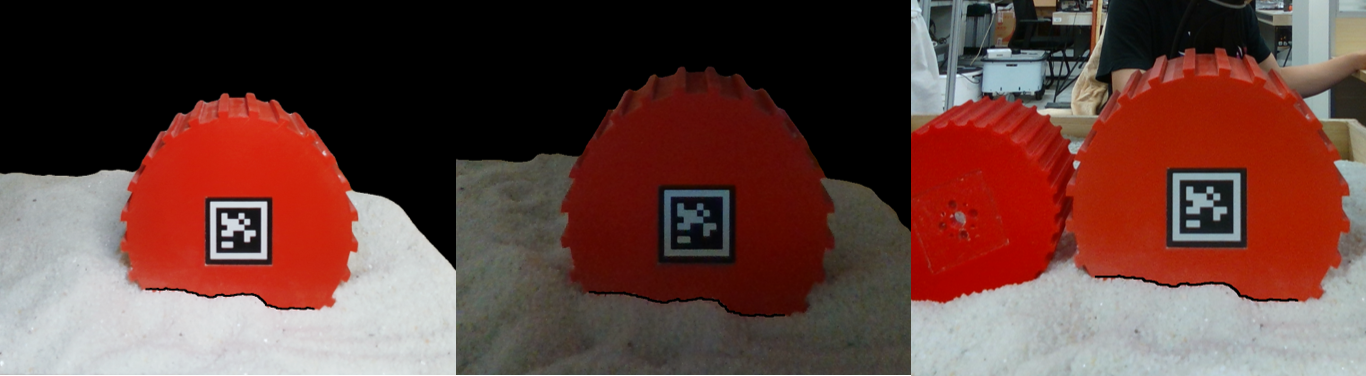} 
\caption{Recognition of wheel-soil contact surface under different light conditions and complex environment, we can detect the wheel-soil contact surface in  }
\label{fig:light}
\end{figure}

We use a linear measurement unit to verify the accuracy of this method, and it turns out that our error range is within $5\%$.
We present a comprehensive set of experiments to validate our approach.  
Experiments were performed under different conditions including non-flat terrains, variable illumination conditions, and different terrain.
It proves that the approach we propose can detect wheel-soil boundary accurately under various illumination conditions.
This means that our detection algorithm is less affected by the degree of illumination, has better robustness and practicability.
Meantime, Fig.~\ref{fig:light} shows that terrain type and complex background can not change to effect of detection, this proves that our detection algorithm has better robustness when rover driving in a complicated environment.

\section{Wheeled Bevameter Experiments}
\label{sec:bevameter-experiment}

When a rover driving on unstructured terrain, it can recognize terrain type by using semantic segmentation or other promote prediction methods, but more concrete parameters such as sinkage and slip ratio require on-position measurement.
On-position measurement often means greater danger because the robot may not escape from soft soil.
But sinkage and slip ratio plays a critical role in mobility prediction.
Our wheeled bevameter can measure sinkage and slip ratio accurately without a robot moving on dangerous terrain.

\subsection{Predict Sinkage by Pressure-sinkage Experiment}
\label{sub:Sinkage_exp}

Sinkage is a significant risk for robots navigating on soft ground. 
The mission will fail if the sinkage is too bigger, and the robot may be damaged.
Traditional sinkage prediction, on the other hand, relies solely on empirical data based on the soil material, which is prone to large mistakes because sinkage is affected by a variety of factors such as moisture.
But if a robot carrying a wheeled bevameter, it can measure sinkage accurately without danger.

We use a wheeled bevameter to apply force to soft terrain in order to determine the connection between normal load and sinkage.
In each experiment, we record the current signals of each servo, the timestamp, data from all channels of the F/T sensor, and sinkage.
To reduce the inaccuracy produced by changing the particle spacing, we repeat the experiment three times and average the sinkage.
Meanwhile, we tested the approach on three granular mediums to verify its generalization.

\begin{figure}[ht]
    \setlength{\abovecaptionskip}{0.cm}
    \centering
    \includegraphics[width=\columnwidth]{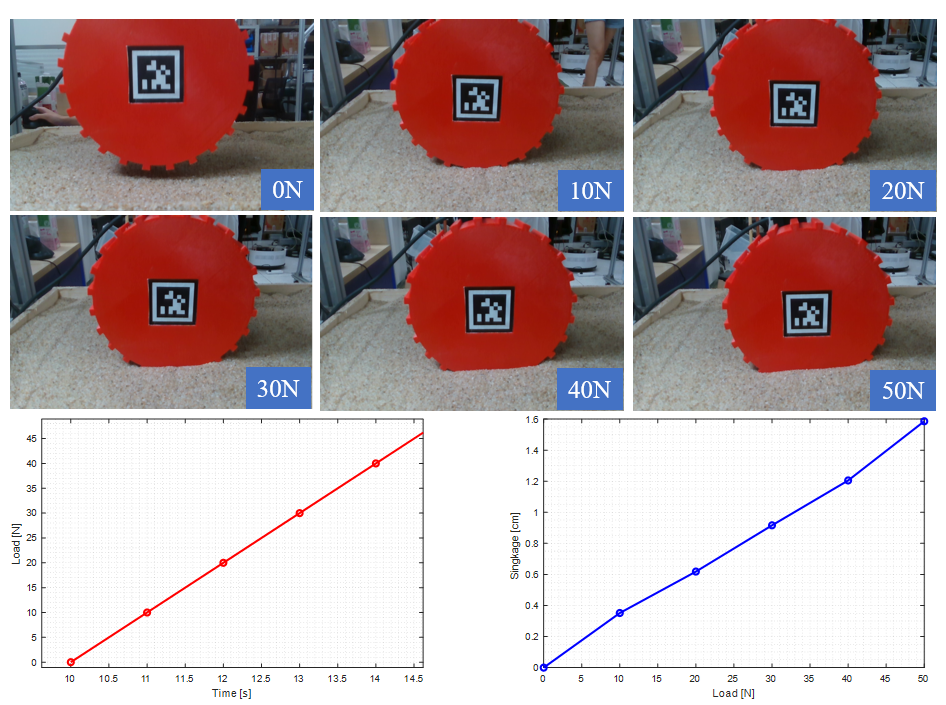}
    \caption{Example of the sinkage-pressure experiment.
    The figures on the upper is a screenshot of the experiment under different pressure, the figure in the lower right corner means the load varies with respect to time, and the figure in the lower right means the sinakge varies with respect to load. 
    }
    \label{fig:raw_data_sinkage}
\end{figure}
The procedure we followed can be briefly described by following step:

\begin{enumerate}

    \item	Control shoulder joint and elbow joint to exert normal force to the soil surface. 
    We can acquire the data of the force sensor at $100Hz$ frequency and calculate the normal load by the transform matrix.

    \item	When the normal load is multiples of $5N$ from $20N$ to $70N$, we measure the sinkage, entry angle, and exit angle with the method in the last section.

    \item After every experiment, we loose and flatten the sand to make the surface in a similar condition.

\end{enumerate}



\begin{figure}[ht] 
\centering 
\includegraphics[width=\columnwidth]{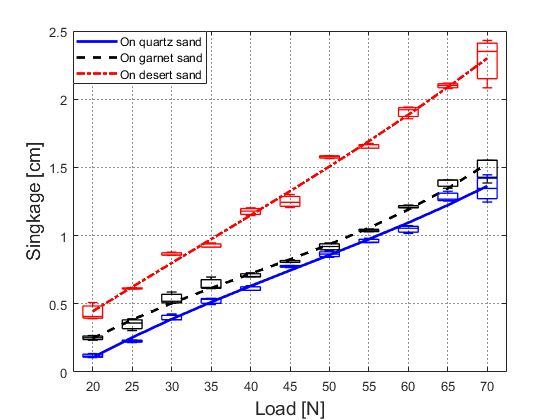} 
\caption{Sinkage of the wheeled bevameter on three soils.}
\label{fig:sinkage-experiment-big} 
\end{figure}

Fig.\ref{fig:raw_data_sinkage} shows an example that the experiment is carried on a wheel-soil interaction platform with a vision collect system.
As can be seen from Fig.\ref{fig:sinkage-experiment-big}, the lines labeled quartz sand, garnet sand, and desert sand represent the change of wheel sinkage with a normal load on the quartz sand, garnet sand, and desert sand, respectively.
Under the same load, the wheel sinkage will change on different soft sand because three types of deformable terrain have different terramicanical properties. 
But it is no doubt that the sinkage will increase as the load increases.
In the experiment where the soft medium is desert sand, the sinakge changes most significantly with normal load, followed by garnet sand, and quartz sand is the least obvious.
This is determined by the properties of the three soft media. 
Desert sand particles have a small density and large gaps, while quartz sand has large and small gaps, while garnet sand is somewhere in between.

Based on the above data, we can regress a quadratic function, the expression is as following formula:
\begin{equation}
    s_k = a{f_N}^2+bf_N+c
    \label{eq:sinkage}
\end{equation}
$f_N,s_k$ denote normal force and sinkage, $a, b $,and $c$ are respectively coefficients.
In practical application, we can directly carry out the experiment under the load of 70N, and record sinkage when the load is multiples of 5N, then the above function can also be obtained.
When the robot crosses a steep slope or the robot's load changes more frequently, this model can predict sinkage of the robot.
And we can increase the mobility of the robot by controlling the center of mass of our robot to change the force of the robot's wheels.
When the force sensor of the robot body is damaged or malfunctions, we can also judge status by identifying sinkage of the wheel.

\subsection{Predict Slip Ratio by Shear Experiment}
\label{sub:Shear_exp}

High levels of slip can be observed on some deformable terrains, which can lead to significant slow down of the vehicle, inability to reach its predefined goals, or, in the worst case, getting stuck without the possibility of recovery.
Slip ratio is defined as the difference between theoretical and practical speed. After normalize it is given by:
{\small
\begin{equation}
\label{eq:slip_ratio }
    s=
    \begin{cases}
    (rw-v)/rw \quad (if |rw|>|v|: driving) \\
    (v-rw)/v \quad \quad  (if |v|>|rw|: braking)     \end{cases}
\end{equation}}
where $w$ is the rotation speed of the wheel, $r$ is the radius and $v$ is the rover's velocity.
A positive slip ratio implies that the rover is traveling slower than commanded, and a negative slip ratio means the rover is traveling faster than commanded.
When the slip ratio is 1, the rover is completely stuck in the rough terrain.
If the slip ratio is too large, the robot will dig the soil and make it sink deeper, making it impossible to get out.
At the same time, slippage will reduce the odometric accuracy, which will lead to inaccurate positioning and planning.
So when the robot is driving on soft soil, the slip ratio measurement and prediction are very important.

\begin{figure}[ht]

\setlength{\abovecaptionskip}{0.cm}
\centering 
\includegraphics[width=\columnwidth]{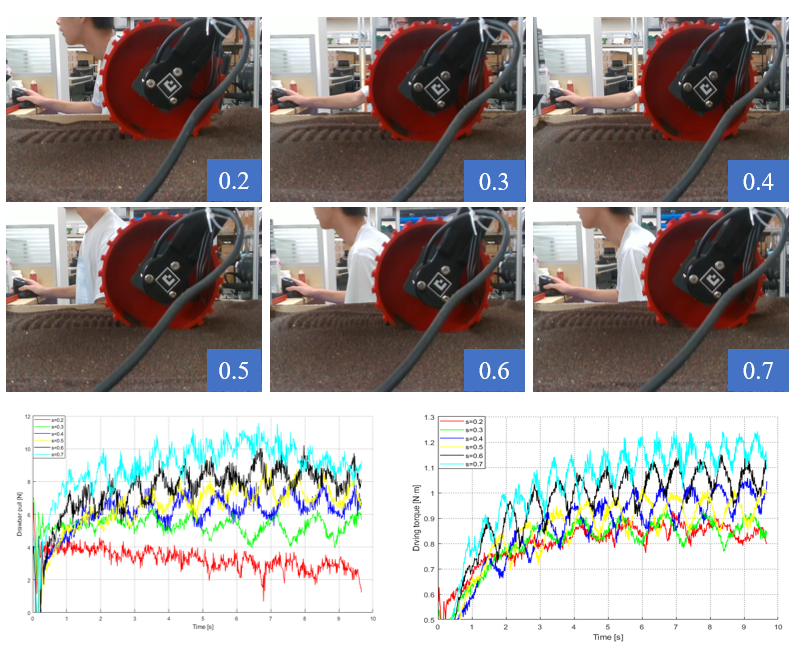} 
\caption{Example of the shear experiment, figure in the upper are some pictures taken during the experiment on garnet sand, the figures in the lower are drawbar pull and driving torque varies with time in different slip ratios.}
\label{fig:raw_data_slip} 
\end{figure}

There are many factors that affect the slippage, such as slope angle and particle density.
Ref.\cite{rothrock2016spoc,cunningham2017locally} mainly depended on terrain classification or empirical models to predict slip ratio.
Obviously, for that environment without available dataset, the classification and prediction will be inaccurate. 
And sometimes those similar soft surfaces would have different slip ratios due to various factors.
Ref.\cite{ishigami2008terramechanics} verified that slip ratio is closely related to drawbar pull and driving torque, so we use the wheeled bevameter platform to find the relationship between them. 
The steps taken are:
\begin{enumerate}
    \item Exert $35N$ normal load on the wheel of wheeled bevameter because our robot is $14kg$ in total, and each wheel will take $35N$ on average.
    \item Many planetary rovers drive at a very slow speed to make sure security, so we control the linear motion moves forward along a horizontal direction at 0.01m/s.
    Meanwhile, we control the wheel rotation speed to keep the slip ratio at the value we want in each experiment.
    \item Experiment with different slip ratio changing from 0.1 to 0.8 and after each experiment, we will loose and flatten the sand to ensure the surface keep similar condition.
    
\end{enumerate}

\begin{figure}[ht]  
\setlength{\abovecaptionskip}{-4mm}
\centering  
 \includegraphics[width=\columnwidth]{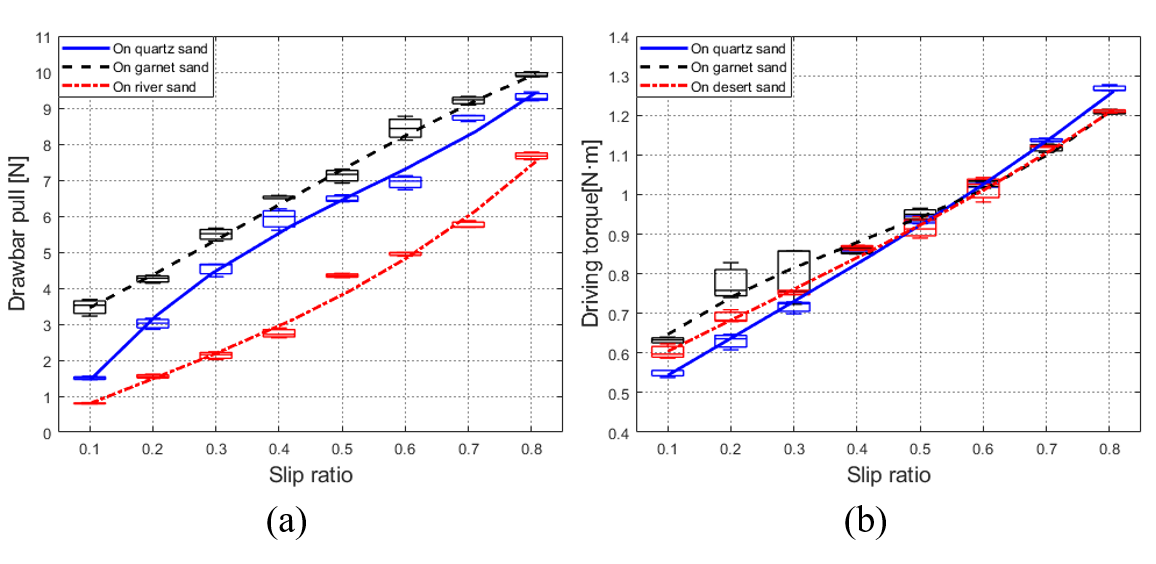}  
\caption{Drawbar pull and driving ratio varies with respect to the time of the wheeled bevameter on three soils.}  
\label{fig:slip-experiment-drawbar-pull}  
\end{figure}

We collect the current signal of servos, timestamp, data in all channels of the F/T sensor, driving torque, and drawbar pull (Fig.\ref{fig:raw_data_slip}).
With the slip ratio increasing, the rutting becomes denser, rotation speed increasing caused this phenomenon.
For each slip ratio, we repeat three times and get the average to reduce error.
After experiments on three types of soil, we get the Fig.\ref{fig:slip-experiment-drawbar-pull}, the lines labeled quartz sand, garnet sand, and desert sand represent the change of drawbar pull or driving torque with respect to slip ratio on the quartz sand, garnet sand, and desert sand, respectively.
As shown in the two figures, the drawbar pull and driving torque of the rover will increase with the slip ratio increasing.
And it is obvious that in different soil, the slope of the curve of drawbar pull with slip ratio is also different, 
this is because different sand granule has very big difference different mechanical properties.

Based on the above data, we can regress a quadratic function, the expression is as follow formula:
\begin{equation}     
    s_r = a{f_{DP}}^2+bf_{DP}+c 
\end{equation} 
where $f_{DP}$ is drawbar pull, $s_r$ is slip ratio. a, b, c are unknown coefficients.
In practical application, in order to save time, we can change the slip ratio multiple times in every experiment.
So we can get a set of slip ratios, drawbar pull, and driving torque in each experiment, then the above function can also be obtained.
According to the function model and the force status of the wheels of the robot, we can predict the slip ratio the robot will produce in the terrain ahead,  whether it is in a high slippage area and whether it can pass.

\section{Field Experiments}
\label{sec:Field}
%
In this section, field experiments performed on a $1.2 \times 1.5 m $ area of sandy terrain are detailed to validate our approach.
We will confirm that robots using a wheeled bevameter can predict the sinkage and slip ratio of unknown terrain within error and choose the safest path.

\subsection{Verification Experiment} 
\label{sub:verification}


Since the weight of the robot is $14kg$, we control the normal load at $35N$, which can ensure that the sinkage and slip ratio measured by the wheeled bevameter are directly applied to the traversability analysis of the robot, if forces on the four wheels are not same, we can take the maximum value as the normal load to ensure safety.
For reducing the gap, we carried out 5 times sinkage and slip ratio measurements.
Each time the data was collected in a different area to prevent the soil properties from changing due to the previous measurement, because repeated experiments in the same area may cause the soil to be compacted, thereby affecting the accuracy of the measurement.

\begin{figure}[H]
\centering
\includegraphics[width=\columnwidth]{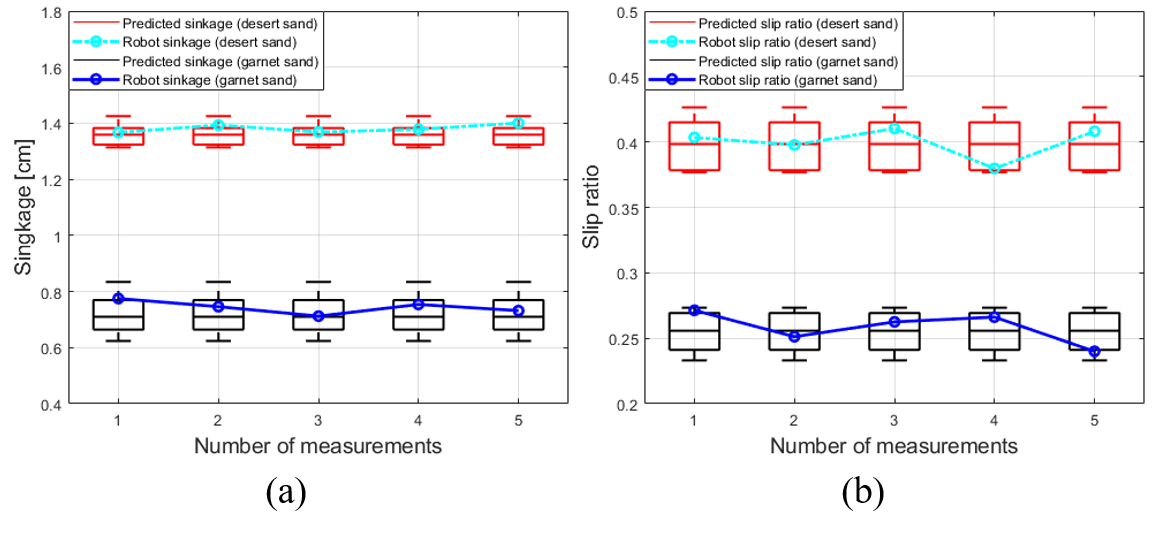}
\caption{The error between the predicted sinkage and slip ratio and the true value on desert and garnet sand.}

\label{fig:ex_sinkage}

\end{figure}

After detecting the unknown terrain using a wheeled bevameter, the robot drives on the soft soil at the target speed and measures its sinkage and slip ratio by a vision system and IMU.
As shown in Fig. \ref{fig:ex_sinkage}(a), the predicted sinkage boxplots represent measured sinkage by wheeled bevameter on the desert sand and garnet sand under $35N$, the robot sinkage lines represent sinkage of the robot on the garnet sand and desert sand respectively. 
In Fig. \ref{fig:ex_sinkage}(b), the predicted slip ratio boxplots represent measured slip ratio by wheeled bevameter on the desert sand and garnet sand, the robot slip ratio lines represent slip ratio of the robot on the desert sand and garnet sand respectively. 
We can find that the sinkage and slip ratio in the actual driving is not much different from predicted by the wheeled bevameter and is always within the predicted value range.
It shows that the wheeled bevameter can achieve an accurate prediction function.

\subsection{Path Selection and Mobility Analysis}
\label{sub:Path_sel}

According to Fig.\ref{fig:whole_robot_model} and Fig.\ref{fig:ex_sinkage}, when the robot moves on different terrains, sinkage and the slip ratio may vary greatly depending on the terrain.
They are the two parameters that have the greatest impact on mobility in soft soil mechanics.
So we can classify the soft terrain and choose the safer path according to them.

We can calculate the traversability score $T$ based on sinkage and slip ratio information.
According to the constraint of robots such as driving capability, wheel size, we design threshold values
$d_{max},s_{max}$ as the max sinkage and slip ratio that the robots can accept, $d_{min},s_{min}$ means means the safety boundary.
The traversability on the soft terrain is:
\small{
\begin{equation}
T =
\begin{cases}
0   & s>s_{max} \ or \  d>d_{max} \\
1 & s<s_{min} \ and \  d<d_{min} \\
min(1-\omega_1\frac{s}{s_{max}}-\omega_2\frac{d}{d_{max}}) & otherwise
\end{cases}
\end{equation}}  
where the $\omega_1$ and $\omega_2$ are the wights, which imply the preference for traversability to the sinkage and slip ratio. 
In our case, we use 0.5 and 0.5, respectively.


So by evaluating the slip ratio and sinkage on the unknown soft terrain, the robot can choose a safer path, predict traversability.

\section{Conclusions}
\label{sec:Conclusion}

In this paper, we proposed a robust and accurate method to estimate contact parameters between wheel and soil.
We designed an unique articulated wheeled bevameter, we can get accurate sinkage and slip ratio while the robot need not travel on the unknown and dangerous terrain using this instrument, which is better than the current in-situ perception and remote perception methods.
Moreover, we experimented to verify that the robot could analyze mobility by comparing sinkage and slip ratio and choose a safer path.
For the future, we will take slip ratio and sinkage on more terrains into consideration and combine elevation map to create a travesability map.
Last but not least, we can even change the terrain with a wheeled bevameter to create a more suitable path.




\bibliographystyle{IEEEtran}
\bibliography{main.bib}
\end{document}